# The AGINAO Self-Programming Engine

**Wojciech Skaba**                                    WOJCIECH.SKABA@AGINAO.COM
*Aginao*
*Trubadurow 11*
*Gdansk, 80205, Poland*

**Editors:** Kristinn R. Thórisson, Eric Nivel, Ricardo Sanz

## Abstract

The AGINAO is a project to create a human-level artificial general intelligence system (HL AGI) embodied in the Aldebaran Robotics' NAO humanoid robot. The dynamical and open-ended cognitive engine of the robot is represented by an embedded and multi-threaded control program, that is self-crafted rather than hand-crafted, and is executed on a simulated Universal Turing Machine (UTM). The actual structure of the cognitive engine emerges as a result of placing the robot in a natural preschool-like environment and running a core start-up system that executes self-programming of the cognitive layer on top of the core layer. The data from the robot's sensory devices supplies the training samples for the machine learning methods, while the commands sent to actuators enable testing hypotheses and getting a feedback. The individual self-created subroutines are supposed to reflect the patterns and concepts of the real world, while the overall program structure reflects the spatial and temporal hierarchy of the world dependencies. This paper focuses on the details of the self-programming approach, limiting the discussion of the applied cognitive architecture to a necessary minimum.

## 1. Introduction

By *self-programming* we mean a computer system that is characterized by at least the following key features:

- the code is "executed" as a program on a (possibly virtual) processor having the flexibility of a UTM or a close-to-universal Turing Machine (TM)

- the ordering of the instructions of the program code, of a possibly fixed instruction set, is a result of an automatic process, rather than prescribed by a human designer

Many other terms have been used to mean similar approaches in the past. *Automatic programming*—once regarded as automatic computer-program generation—now is more understood as generative programming, an assembly of components with template tools, the process that barely meets even the second criterion (Batory, 2004). In e*volutionary programming,* the program structure is fixed, while the parameters may evolve (De Jong, 2006). *Genetic programming* involves modifications of the program code but imposes restrictions on the overall program structure to be represented by a tree (Koza, 1992). The *universal search* and derived methods are the closest to our understanding of self-programming, nevertheless the papers on the topic rarely deal with the obvious question of combinatorial explosion in an effective manner (Schaul and Schmidhuber, 2010). Few if any TM-based attempts find the underlying self-programming task as designing a real-time embedded system processing huge amounts of data, that is unknown a priori, volatile and affects the program outcome substantially. The latter to mean that a piece of code must be rerun multiple times to evaluate its usability. We do not reserve, however, the right to restrict the extent of the meaning of the term self-programming, providing the above definition to focus attention only.

The choice of TM-based approach is obvious for one reason: it is the most universal method of creating a dynamic non-linear system, capable of running any algorithm, according to Church-Turing thesis. The universality is given at the price of system stability, as the question of executing an illegal instruction, infinitely looping, accessing data out of scope, or—in general— the halting problem arises. No system is capable of being a self-programmable from scratch. There must be a handcrafted core layer, and only on top of that comes the self-programming layer. The upper layer need not to know anything about its underlying core. It may be a simulated UTM running in a virtual memory, while the core is a rather restricted system, far from being universal, thus stable. On the other hand, however, there is no limit for a future AGI system, running in the upper layer, to learn its core and even rewrite itself (Schmidhuber, 2006).

## 1.1   Cognitive Architecture

The AGINAO cognitive architecture was first introduced by Skaba (2011). The cognitive architecture is built around the notion of a pattern and a concept. Here, the idea of a pattern is intended to mean much more than a simple graphical icon; it is any regularity that might be detected in the world. For example, a simple pattern is the letter **T**, being a purely spatial pattern, consisting of a horizontal bar and a vertical bar, both arranged in a specific order. The bars themselves being the patterns of some lower level ingredients (pixels?), and the letter T being a candidate for a higher level pattern, like a word. It is an open question, however, whether the letter T consists of two bars, or—let's say—two horizontal half-bars and one vertical bar, or some completely different ingredients, or even coexisting all alternative definitions of the same pattern?

A pattern need not be spatial but may be temporal, like an output produced by a pace maker. It may also be a spatial-temporal, like any object moving around a visual scene. Not only episodic but also procedural memory is an example of a regularity, like an order of the actions to be executed. A regularity need not be preexistent in the environment, for it may also be envisioned by the cognitive system, by constructing a deliberated pattern. Possibly, the approach in which patterns are envisioned and evaluated for their applicability may be more effective than an attempt to directly reverse-engineer the incoming data.

One of the simplest regularities one might imagine is a detector of a threshold-exceeding input signal. It doesn't look like a pattern at all, and it is unclear how it might be applied. In terms of program complexity, however, such a detector is a short piece of code, very likely to come forth even as a result of a random code generation. What is more, it may be applied in different locations of the cognitive engine and in different contexts. We may expect these types of regularities, having no clear functionality and no names, to become very common.

A patternist philosophy of mind is thoroughly discussed by Goertzel (2006). According to Goertzel, the mind and world are themselves nothing but pattern—patterns among patterns, patterns within patterns. Hawkins (2006) was among the first who have focused on the temporal aspects of the patterns. He directed his attention to the following ideas:

- most real-world environments have both temporal and spatial structure, and a single algorithm to discover these patterns should take both aspects into account

- better results can be achieved, if the processing is conducted simultaneously at all levels of the hierarchy

Even if a pattern is a purely spatial one, we often use a temporal language to describe its properties. For example, we would say that the letter E **comes** before T in word *letter*. The way we learn the letters, the way fovea saccades over a static letter, the way we learn how to write them, all these actions engage time.

Once the cognitive engine has created an internal representation of a pattern or regularity, we call it a *concept*. A concept may sometimes represent a fake pattern, as a result of a false belief, and will be discarded eventually. In most cases, a concept will be represented by a piece of code, a separate subroutine that may be launched independently of the other concepts, or even simultaneously many instances of the same concept code, possibly processing different data. In a more general sense, the notion of a 'concept' could be attributed to a number of separate concepts acting concurrently. We wouldn't use that name, however, unless a higher level concept has been formed and translated to a piece of code.

Figure 1 depicts the basic scheme of a concept. Each concept has one or more inputs, and a single output. The inputs link from the outputs of the other concepts. The output links to the inputs of the other concepts. In general, a new pattern is formed out of other patterns in two steps. First, there are ingredients, the patterns of which a new pattern may be formed. Second, there is some internal processing. For example, a letter T consists of a horizontal bar pattern and a vertical bar pattern ingredients. It is, however, the internal processing that tells whether the ingredients are arranged in a proper configuration. In general, a pattern is any set of patterns in some mutual relation, spatial and/or temporal, while a concept is the way the cognitive system represents the patterns.

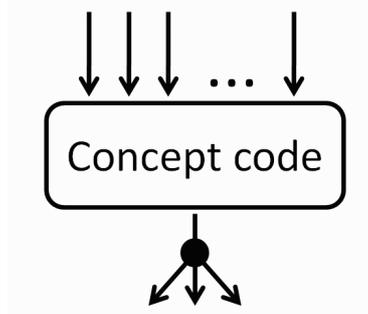

Figure 1. Basic scheme of a concept.

A question arises on the distinction between a concept template (the program code and I/O structure) and its instantiation, later called a *runtime* for it is an executable of a concept that is running. It is the runtime that expresses the idea that a given concept is currently perceived or imagined. In some cases, new concepts are formed only if multiple runtimes (executables) of a certain concept coexist. A good example would be to consider a collection of natural leaves. If runtimes of a given leaf-concept have the property of having the same creation time (the time they were perceived), they could form a concept of a tree. If, however, the same number of runtimes of the same leaf-concept is differing in the creation time, it is subject to supporting a concept of a falling (single) leaf. It is these considerations that justify the necessity of the mentioned above simultaneous processing.

Since a concept may be created out of two or more instantiations of another concept, it means that it is quite legal for a concept-output to link to two or more inputs of a newly created concept, provided that—when the processing is conducted—the different inputs are represented by different runtimes. Two different concepts may also share the same input mappings, provided they differ in the internal processing code, like in the above example.

Different world-patterns feature different complexity. They are arranged in hierarchies and according to some mutual dependencies. We would like the concepts for the simplest patterns to emerge first, then—on top of them—the more sophisticated ones. Happily, we have a good candidate for a measure of complexity, the Kolmogorov complexity. Instead, however, of trying to measure the Kolmogorov complexity of each individual pattern—a task in general case incomputable—we could test the individual programs in order of their complexity, taking into account both the program length and its execution time, the latter to matter even more. Then, the patterns to be named the simple ones would be those detectable by simple program. As will be shown below, not all possible programs match the criteria of functioning as a concept, but some do. A similar approach was presented by Schmidhuber (2004).

Some authors doubt if the presented above idea of creating a hierarchy of concepts representing spatial-temporal patterns would have sufficient power to eventually scale-up to an advanced AGI system. Skipping the discussion here, we would refer to Goertzel (2011).

## 1.2 NAO Humanoid Robot Embodiment

Advanced AGI system may only be achieved, if we let the cognitive engine act as an agent in the natural environment. One possible approach is the embodiment of the control program in a physical humanoid robot. Though many teams have succeeded in constructing their own hardware platforms, a more straightforward approach for a cognitive scientist is the application of one of the commercially available products, like—in our case—the NAO from Aldebaran Robotics.

One question that immediately arises and, as so far, remains unanswered is: do the NAOs sensors and actuators meet the embodiment requirements for achieving a HL AGI? Taking, for example, the visual system into account. The robot's camera resolution is 1280x960 pixels, much less than that of the human eye. Would this restriction affect the possibility of creating an AGI system substantially, or would it only have a minor impact? One must observe, however, that there are people who have been blind since birth and have developed regular intellectual skills in the very meaning of the sense of general intelligence, with the only exception that they do not deal with visual images. Would then vision processing be required for achieving

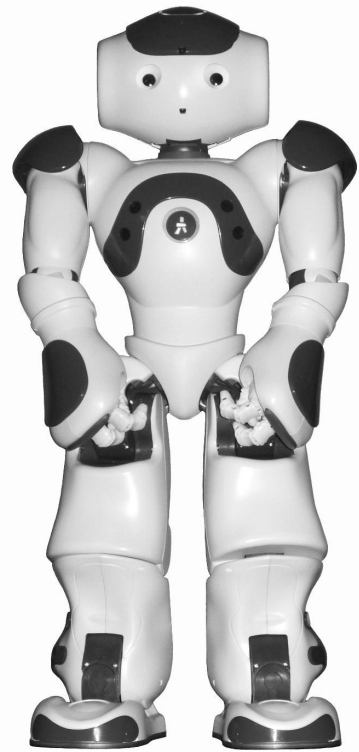

the HL intelligence at all? We will leave that question unanswered.

The cognitive engine embodiment adds another requirement on the construction of the whole system: the necessity to connect it to the outer world. The concepts—as defined above—cannot be connected directly to sensors and actuators, since the inputs link from other concept's outputs and the outputs may link to other concept's inputs. The problem may be resolved by adding special purpose predefined atomic concepts that would act as individual sensors and actuators. These concepts would be indistinguishable from the regular ones by the cognitive system, with the only exception they would be non-removable and would be lacking some properties.

It is the atomic sensory-concepts that act as the most basic patterns for the construction of the more sophisticated ones, for otherwise the cognitive engine wouldn't find the ingredients for the construction of the first new patterns. It may be said that the cognitive system is grounded in atomic sensory-concepts. The atomic actuator-concepts, on the other hand, act as terminal concepts, having no output (as a concept), and thus not participating in the construction of other concepts.

To make the further discussion more comprehensible, the following is a brief presentation of the robot's hardware. The NAO is controlled by INTEL ATOM 1.6 GHz (formerly AMD Geode) running LINUX. Robot is equipped with 2 color cameras—not to be used concurrently—1280x960 pixels max each, 4 microphones, 2 speakers, tactile/force sensors, gyro, accelerometer, dozens of joints with motors and sensors controlling the head, the arms and the legs. The robot may be connected via WiFi or Ethernet. Since the power of the robot's CPU seems insufficient for the planned cognitive task, and occasionally up to 90% of the computational resources are consumed by robot's internal processing, all the cognitive-engine computations are performed on a separate host system, while the robot merely transfers sensory data and executes the commands received for the actuators.

## 2. Self-Programming Engine

The very idea of self-programming may interfere with the reader's beliefs on the theoretical possibility of attributing a human-like creativity and intelligent activity to an inanimate machine. On the other hand, do we really need the self-programming at all? Why cannot we just encode the cognitive engine by hand? The answer is that we have already tried and failed, and the cause of the failure—we believe—is that the number of hypotheses (i.e. programs) to be created and tested is beyond the capabilities of a human programmer.

Let us consider the following thought experiment as an informal proof on the possibility of machine creativity. The prime (integer) factorization problem is believed to have no polynomial-time solution, but algorithms better than brute-force search do exist. As of 2012, the *general number field sieve* is the winner. Now imagine that a better but unknown algorithm exists, one possibly to be discovered as a result of an exceptional intellectual creativity. Then, the algorithm may also be found with the universal search, provided we have a fitness function, which is simply the multiplication of the divisors to check the algorithm's correctness.

As for a generalization of the above approach to a more advanced AGI system, we encounter the following crucial questions:

- universal search has exponential-time complexity and—as for the expected size of the AGI core—we do need a way to expedite the search

- the fitness function for general intelligence is unknown, or—at least—it is not as straightforward as in the case of integer factorization

- it seems rather unlikely to immediately generate a complete mind with all its experiences and memories; we would rather focus on making a core system to be embodied in a robot, one that would start building the target mind from experience. What follows, no matter how fast our core system is, we have to test each candidate core for—let's say—three years, until we get any reasonable results. Some authors attempted to overcome this limitation with a simulation world (Heljakka et al., 2007).

## 2.1 Foundations of the Computational Model

Taking into account the discussed above assumptions and restrictions, the following basic model of computation has been proposed:

- at the lowest level, there is a simulated Turing Machine, later also called a virtual machine (VM), with a predefined and fixed (though customizable) instruction set, capable of running a virtually unlimited number of concurrent threads. A single VM model is shared by all concepts.

- out of the VM instructions, in a fully automatic process called *heuristic-search in program-space*, the basic building blocks of the cognitive engine are composed, the tiny programs named codelets, ones that—accompanied with I/O structure and other information—become the concepts. The codelets must be big enough to match the minimum criteria of a concept structure, and small enough to be manageable. The former to mean that the code to match our assumed concept structure must at least read some input, do some internal processing and generate an output. The latter to mean that, as the codelets are created by a heuristic-search on top of a random generation of the codelets, the longer pieces of code are less likely to perform any reasonable task.

- the generated concepts are stored as a repository of executable programs arranged in a hierarchy of dependencies, with atomic sensory concepts at the bottom of the hierarchy, atomic actuator concepts as the terminals, and self-generated concepts on many layers in between. Figure 2 depicts a sample hierarchy. Both the creation and the arrangement of the concepts are results of a fully automatic process that will be discussed in detail later. For the time being, it is important to notice that the concept network is not a neural network, but a list of potentially executable programs and unidirectional connections. At any time, multiple executables (runtimes) of the same concept, possibly processing different data, may coexist. The connections specify the order of execution and the data flow. The hierarchy of the concepts is a dynamic and open-ended structure. A concept as a whole, or an individual link between pre-existing concepts, may be added to or removed from the hierarchy at any time. A concept/link removal may occasionally cause an avalanche of removals of many descendent concepts and connections, a phenomenon that is observed on experiments. Any input may be connected to many concepts. The last input link of a concept that is removed invalidates the concept and causes the concept deletion and all the aforementioned consequences.

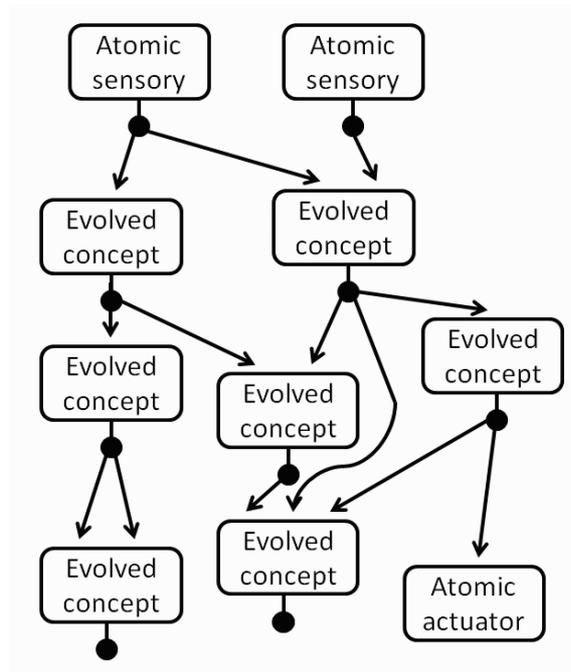

Figure 2. Sample hierarchy of concepts.

- the execution is conducted as follows. At any time, a concept may have its runtime copy (possibly one of many concurrent but independent) executed by the VM in operational memory as a single thread. Once the execution is completed (`TERMINATED`), the runtime refers to its concept structure to get a list of the descendents, to determine where to pass the result to, and effectively which concept(s) to launch runtimes as the next. For each selected descendent, a new independent runtime thread is created. At the same time, the output from the completed execution becomes the input of all the launched descendent threads. It is also very likely that any launching of new threads is abandoned for many reasons discussed below, e.g., lack of computational resources.

- the selection of the descendents to pass the execution to is dependent on the probabilities computed from the values of all possible next actions, computed with a TD-learning algorithm. Thus, with each completed execution, the reinforcement-learning (RL) update is performed.

- not only a selection of the next action from a pool of the available ones, but also an addition of a new action—namely creation of a new outgoing link—may be performed in the `TERMINATED` state. Unlike common convention, we would use here the name *exploration* for adding a new link-action, and the name *exploitation* for passing execution to any of the pre-existing descendents, not only the most greedy ones. In fact, we do not even refer to the distinction between greedy and non-greedy. The exploration step is usually preceded by a construction of a new concept or selection of an existing one. Once the creation/selection is completed, the new link is added. If the new concept contains more than one input, the other inputs may be connected to the outputs of other concepts.

Figure 2. Sample hierarchy of concepts.

A thread may also be launched by an atomic sensory-concepts. Since atomic concepts do not perform any internal processing, but look like regular concepts, an atomic sensory-concept runtime is created directly in the `TERMINATED` state, signaling execution completion, and delivering an output value representing the value of a related sensor. In case of the visual system, there is a single atomic-concept representing a pixel that is shared by all pixels, with the output consisting of (row,col) coordinates and color in (Y,U,V) space. It is quite likely that 100.000+ independent visual pixel threads are launched within one second, ones differing not only with (row,col,Y,U,V) parameters but also with the creation real-time.

A hidden mechanism decides when a sensory concept thread is launched. In case of the visual system, for example, a runtime is created when the difference between the former and the current value of the pixel's luminance (Y) exceeds a predefined threshold. In fact, the sensory threads are the very first ever to arise and the ones that keep the whole cognitive engine running.

One might notice that a computational model, as presented above, might very quickly encounter the combinatorial explosion problem. If every thread is capable of launching a number of new threads, and the total number of concepts capable of launching threads is growing in time, we would very quickly run out of the computational resources. To deal with that problem, each thread is assigned a priority, a computational-resources limit and expiration time, differing from thread to thread, according to some rules that are discussed in the section on Artificial Economics. Once a thread is created, it is placed in a priority queue first, awaiting for being serviced by the scheduler, unless the expiration time passes first, in which case the thread is discarded as if it has never existed. Once executed, it may consume the VM resources (clock units) only until the assigned resources limit, for otherwise it would be discarded. This feature is a protection against infinitely looping and favors faster programs, as having more chances to complete execution within the assigned resources. If computation is completed, the thread may launch new threads, or create new concepts, but only within the limits of the remaining resources, and provided the expiration time has not passed.

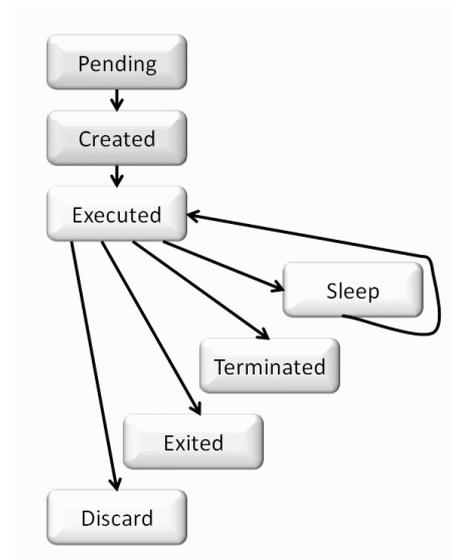

Figure 3. Runtime life cycle.

## 2.2 Runtime Life Cycle

A runtime, during its life cycle, may be in one of the seven distinct states, named: `PENDING`, `CREATED`, `EXECUTED`, `SLEEP`, `TERMINATED`, `EXITED` and `DISCARD`. Figure 3 shows the states and the possible transitions between them. Some runtimes, like the aforementioned sensory threads, are governed by slightly different rules.

- the first possible state is the `PENDING` state, the case when a thread has already been created, but for some reasons is yet not ready for execution. The most common case it occurs is when a multi-input concept runtime was requested, but one or more of the inputs are still missing. According to experimental statistics, 70% of the `PENDING` runtimes are discarded before transiting to the next state, due to passing the expiration time. A `PENDING` state, if discarded, does not transit to a "discarded" state, but is discarded immediately, for there is no `DISCARDED` state, but only an imperative request state `DISCARD`.

- the `CREATED` runtime is one that is ready for execution but awaiting in the priority queue for being serviced. In case of a single input concept, a runtime is created directly in the `CREATED` state, skipping the `PENDING` state, for a new thread may only be requested if at least one input is available, and a single-input-concept runtime need not be pending for any other input(s). A `CREATED` runtime may also timeout. The collection of runtimes on top of the priority queue may be said to reflect the attentional focus of the whole system.

- the `EXECUTED` state is the state of running the code on the VM. The `EXECUTED` runtime may also be discarded before the execution completes. The most common cases are: running out of resources or an attempt to execute an illegal instruction, like accessing data out of scope.

- a running code may go a-`SLEEP`, if the temporal-delay instruction (`WAIT`) is encountered. These instructions are very important for managing the temporal patterns. Since minimal gaps between real world events are counted in tens of milliseconds, compared to instruction execution times counted in nanoseconds, instead of looping for quite a long time and consuming the VM resources, a thread is suspended and placed in a temporal priority-queue. Once the designated delay-time passes, unless the expiration time has passed first, the execution is resumed. Effectively, a SLEEP-ing thread is in an idle state and does not utilize the VM processor.

- the execution may finish in `EXITED` or `TERMINATED` state, the former a result of encountering the `EXIT` instruction, the latter a result of encountering the `RET` instruction (see Appendix A for reference). The `EXITED` state means a successful termination with a negative outcome, and may occur—for example—in case like the discussed above problem of detecting the letter **T**, if a horizontal and a vertical bars (as inputs) are not properly arranged. Effectively, any further processing of the thread, like passing data to other concepts, is discarded. The state, however, is not regarded as a runtime error.

- the `TERMINATED` state denotes a successful completion of the execution and a positive outcome, especially a detection of the underlying pattern, and a possibility of launching new threads (exploitation) and adding new concepts (exploration). As has been mentioned before, the sensory concepts are created directly in this state.

- in some cases, a runtime is destined to be discarded, but for some reason must remain in the operational memory, in which case it is marked as `DISCARD` state. This is basically the implementation question, for the physical threads may be locked and some operations must be postponed until unlocking.

## 2.3 Main Execution Loop

Figure 4 presents a detailed diagram of the main execution loop of the cognitive engine:

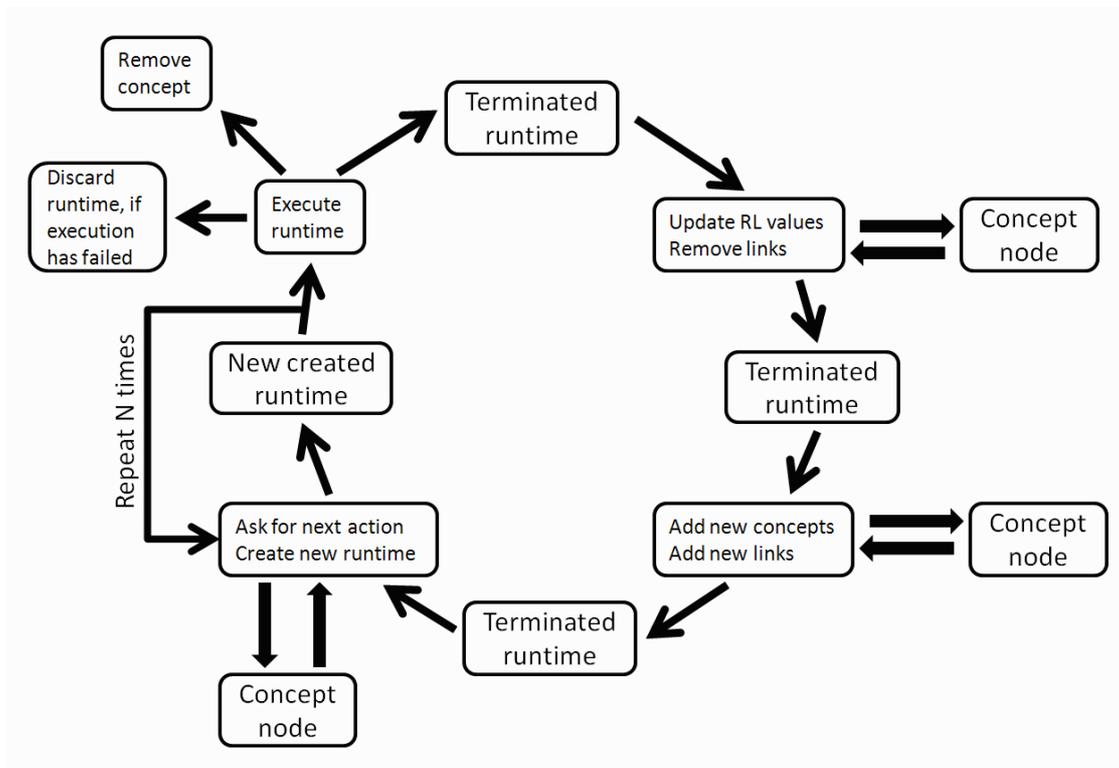

Figure 4. Main execution loop.

It cannot be said which is the first stage on the depicted above Ouroboros-like loop. At the very first start-up of the cognitive engine, the atomic-sensory threads are created: a separate thread for each sensory-event, e.g., for each visual pixel at any change of its luminosity exceeding a predefined threshold. The threshold level—on the other hand—is varying too, in order to adjust the flow of the sensory data to the current system load and the momentary volatility of the input.

All sensory threads are created in the `TERMINATED` state and the processing starts in a stage depicted as the rightmost *Terminated runtime* of the execution loop. From then on, the operation of the cognitive engine is continuous, not excluding the initiation of new sensory-threads. It cannot be said which threads are dominant, since up to a few millions threads are created within each second and—excluding the waiting time in the priority queue—each thread lives for no more than a couple of tens of nanoseconds. A virtual flow of information processing is composed out of tiny threads, ones that [potentially] spawn multiple other threads each, with no any obvious path to be drawn from the roots to the leaves.

We will start the presentation of the main execution loop from the top-most *Terminated runtime* stage (runtime in `TERMINATED` state). First, the reinforcement-learning values, related to the terminated execution, are updated. It must be highlighted, however, that—what is updated—is the values assigned to the action-links of the parent predecessor concept(s), not the concept underlying the terminated runtime.

In the next stage, the output links of the runtime-underlying concept (the concept which program code is executed as runtime) are removed, provided some conditions are met, like in case when the value of an action goes below some level, possibly not below an absolute level, but a level relative to the values of other competing actions. The other possible cause of removal is the case when another concept, the one the link points to, has been deleted, and the link—now connecting to nowhere—is no longer valid. The concept could have been deleted because it was a multi-input concept and one of the other input-links was removed, due to a cause currently unknown and unimportant. If a concept is missing one of its inputs, it is deleted and, consequently, all the links pointing to the deleted concept are removed. Referring to the former paragraph, should the RL value-updates motivate the action-link removal—one from the parent concept to the current runtime-underlying concept—the link removal would not be performed in the current execution loop, but rather the next time the parent concept encounters the link-removal stage; the reason being that the removal decision is not a standalone one, but dependent on the values of the competing actions—a task too computationally intensive to be executed each time the RL values are updated.

In the next stage of the loop, either a new concept is built from scratch (see section 3), or a new concept as a copy of another concept selected from the hierarchy is created, or a concept in the hierarchy is selected without being copied. Following, new link(s) to the created/selected concept are added. One or more input(s) of the newly created/selected concept will be connected to the output of the runtime-underlying concept of the execution loop under consideration, but not necessarily to all of them. If some inputs of the created concept do still have missing links, the creation process will be now suspended. Unlike in the discussed above case of a concept deleted due to input-link removal, a newly created concept is protected against being deleted in such case. It is suspended in a *pending* status (not to be confused with the `PENDING` state of a runtime) until being completed, or until the concept expiration-time passes (not to be confused with expiration-time of a runtime). The exploration process, as discussed here, is a relatively rare event, compared to a more common exploitation (like 1/1000 on average and decreasing), while its frequency is inversely proportional to the combined value of all concept-actions (for a given concept).

A special type of a concept copied from the hierarchy is a copy from a pool of actuator-concepts (the templates). One or more copies of each actuator-concept may be integrated into the concept hierarchy. Unlike the templates, the copies may be deleted. Should more runtimes related to the same actuator be executed concurrently, the conflicts are resolved.

In the next stage of the main execution loop, the new threads are launched, selected with a probability proportional to the values of the action-links. The process may be repeated until all the `TERMINATED` thread resources are exhausted. It is this stage where the main execution loop spawns new threads and eventually aborts.

The discussed earlier priority-queue awaiting, and a possible discarding of a thread due to timeout, is not depicted on the diagram. Once the runtime processing is started, it may finish in one of the three possible ways. First, the runtime may be discarded and the execution loop aborted, as a result of not reaching the `TERMINATED` state. Second, not only the runtime will be discarded, but also the runtime-underlying concept will be removed. This is usually a result of repeatedly executing an illegal instruction. Third, the thread will continue creating new concepts, new links and new threads, until the thread resources are exhausted.

Once a newly created concept is completed, it is automatically integrated into the hierarchy of the concepts, for all its inputs are connected to the outputs of the concepts that have already been in the hierarchy. One may ask: why the concepts are constructed only out of the input candidate-ingredients (concepts) that have an active (`TERMINATED`) runtime at the time of creation? Why cannot a new concept be created out of any randomly selected concepts from the hierarchy? The answer is, it can. Focusing on active concepts only, however, biases the search towards the concepts that happen to coexist as runtime, i.e., are more likely to detect a repeatable pattern.

## 3. Virtual Machine and Heuristic Search

The exploration step and the concept creation procedure involves a construction of a codelet program that is later executed on the virtual machine. This section discusses the details of the implemented virtual machine, instruction set and the heuristic process of program construction.

The decision to simulate a virtual Turing Machine in software, rather than directly execute the machine code of the host processor, is justified by the three following observations:

- a virtual machine with a conceived internal structure and instruction set facilitates greater flexibility of design, for the structure of the VM need not be fixed, and the instruction set may be customized. Changes in the VM design, however, are not supposed to be a result of an automatic process, nor to be very common.

- a VM enables full control over the resources consumption, especially, the code execution times. It also solves the problem of exception handling, should the processor execute an illegal instruction.

- though the speed of execution of the machine code directly would be higher than that of a virtual simulation, both machines—according to the definition of Turing equivalence—differ only by a constant multiplicative factor. What matters in our challenge, however, is exponential rather than multiplicative reduction of the processing time.

### 3.1 VM Design and Internal Data Format

Figure 5 depicts the internal structure of the VM processor and a typical setting of a 2-input concept. Instead of using a simple deterministic single-tape TM machine with a simple transition function, consisting of a short alphabet and a few states, or even an *ultimate reduced instruction set computer* (URISC), we've applied an architecture resembling those of the microprocessors of early 1980s, that could be defined as a sophisticated deterministic multi-tape TM. The point being that programs created of very simple instruction sets are human-unreadable and relatively long if supposed to display any functionality, hence not very likely to be created by chance.

The basic internal data type (word) is integer (`int`), which in current implementation is a 16-bit integer (`int16_t`). The VM may be said to be a 16-bit processor. There are two `int` registers (accumulator and index), and two binary flags (zero and minus). There is also a local static memory with a fixed size. The number of inputs, the output size and the local-memory size are randomly set for each concept individually.

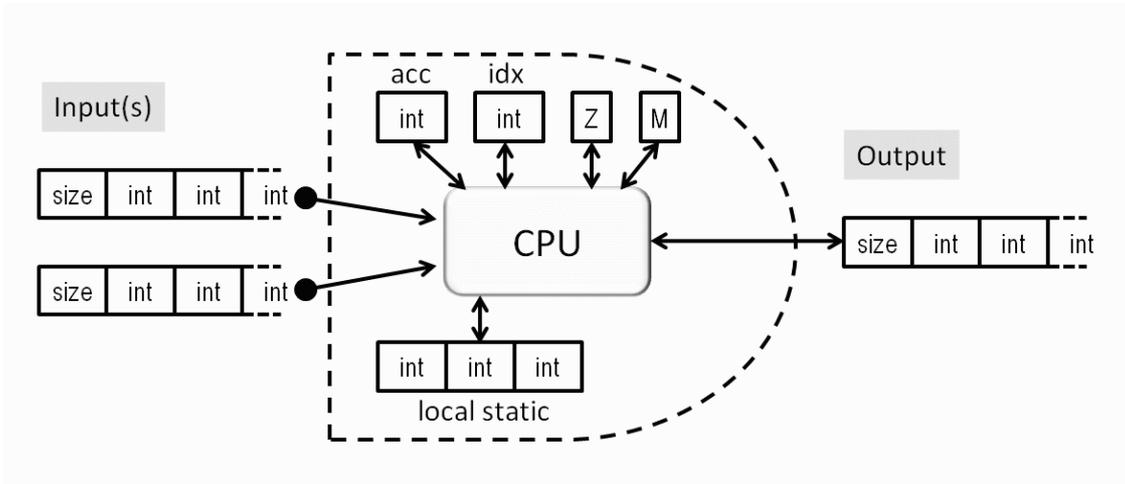

Figure 5. VM internal structure.

Before continuing the presentation of the VM, we have to specify the data format the concepts—or more precisely the runtimes—communicate. There is a uniform data-format, a vector of integers (`int`) of known size (length). The size is stored in cell number 0. This format is similar to Pascal convention. Figure 6 presents a sample visual pixel encoding.

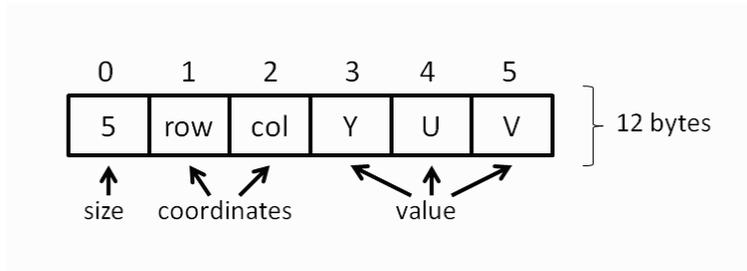

Figure 6. Sample visual pixel vector encoding.

A concept may have one of more inputs, which are always read-only vectors, because a single output of the preceding concept may be connected to many inputs, concurrently, and a race condition could occur. On the other hand, there is no reason to modify the input(s). The concept setting does not restrict the maximum size of the input(s), which—on the other hand—are known at runtime. This property enables, for example, sending a full 5-word pixel vector to a concept processing only the first 2 words (row and column). The internal code has to check that the input vector has the size of at least 2.

Output vector, on the other hand, has a concept-predefined maximum size. The actual size may be smaller and will be stored in the output's size-field. Output vector is a read/write area (from the point of view of a given concept), for it will not be forwarded until the processing is `TERMINATED`, i.e., until the output becomes an input of other concept(s).

Unlike the values of the input and output vectors, which are runtime-specific, the local static memory is a concept-specific read/write area, i.e., shared among all runtimes of a concept. This feature resembles a static variable of a C++ function and is initialized to zero at concept creation. The static memory need not have the size field, for the size is fixed and stored in concept encoding. Local static memory is intended to function as a sort of long-term memory.

Figure 7 present a C++ function equivalent of a concept, where N is the number of inputs:

```
int *prog(const int *src1,..., const int *srcN)
    { static int memory[size];
      // program code here
    }
```

Figure 7. Concept structure C++ equivalent.

## 3.2    Sample programs

Appendix A lists the instruction-set of the VM for reference. Figures 8 shows two sample programs (for an operational example see section: Experimental Results). The line numbers in the first column of each program are counted in bytes.

The first sample program sets the index register to value "2", then stores the value of the third (idx=2) data field of the first input in the accumulator. Next, the value of the accumulator is stored on position 0 of the output, which sets the output size to 1, and the program returns.

The second program moves the value of the first cell of the first input (not the size field but the first data field) to the accumulator, and adds to it the first cell of the second input. The result is appended to the output. If the result of addition is zero, in which case the Z flag is set, the program loops and jumps to the addition again. Otherwise returns. It is quite likely that the program could execute an illegal instruction here, since its outcome depends on the unknown input data. Most likely, the program would exceed the maximum output size, by appending too many words. In some cases (e.g. var2[00]=0), it would also loop infinitely, and would be aborted due to exhaustion of the resources.

```
0000 MOVI IDX,   0002
0003 MOVX A, var1[idx]
0006 SAVI [00],A
0009 RET
```

```
0000 MOV A, var1[00]
0005 ADD A, var2[00]
0010 APPEND, A
0011 JZ 0005
0014 RET
```

Figure 8. Sample programs.

### 3.3 Heuristic-Search in Program-Space

As one might have noticed, the cognitive engine is attempting to build a reliable system by continuously improving the concept network, the process that consists of three concurrently executed phases:

- in the first phase, the candidate concepts are constructed out of the machine code instructions, in a process called *heuristic-search in program-space*. A randomly generated program, even if contains 4-6 instructions only, is not very likely to meet even the simplest criteria for being a valid one. Tricky heuristics may improve the process significantly, i.e., increase the probability of getting a usable codelet.

- once a codelet matches the criteria of the first phase, it is still prone to runtime fatal-errors that are not detectable in the first phase, due to behavior dependent on the input data. This case is well visible in the second of the samples shown in figure 8. If a program frequently encounters fatal errors, the concept will be permanently removed, no matter how would it behave in the third phase.

- the third phase is the time when the concepts are evaluated, based on their individual properties and on their conformity to the overall picture. The reinforcement-learning rules and other tools are applied.

This section focuses on the search in the space of programs and discusses some of the applied heuristics. Readers not interested in the details of codelets' construction may skip this section. The presentation of the heuristics, however, is important for the discussion of the risk of combinatorial explosion and the overall system performance.

The program generator (PG) selects instructions by random, according to some predefined probability distribution that favors instructions that are more likely to be useful. Then the heuristics impose constraints that must be met. There are many obvious constraints that come to ones mind immediately. For example, if a jump instruction is to be inserted in the constructed program, the destination line number should not point to itself, for such a code is completely useless. A more challenging constraint demands that a flag setting instruction must precede a conditional instruction. Likewise, at least one of the program branches must set the output, for otherwise one would get a concept that cannot be used as input of another concept.

The list below presents a selection of total 30+ currently applied heuristics, some being quite tricky. Since typical programs do not exceed 8 instructions, the listed below constraints seem to be quite restrictive:

- only a small subset of the instructions listed in Appendix A would make sense as the first instruction on the program entry (`0000`). Arithmetic operations, flag settings and jumps are then useless, unless some data has been initialized first. Likewise, the last instruction of the program must be `RET`, unless `RET` has already been used elsewhere, in which case the only possible alternative is an unconditional `JMP` backwards or `EXIT`.

- the `A` and `IDX` registers must not be used before being initialized first. A counterpart rule says that, if a register was set, it must be utilized further in the program code, for otherwise the setting would be meaningless.

- flags set *intentionally* must be utilized before the current flag-setting is overwritten by another flag-setting. The `FLAGS` and `CMP` instructions set flags intentionally, while `ADD, SUB, INC` and other instructions set flags only as a side effect, the setting that may be exploited or not. In consequence, it means that—after intentional flag setting—a conditional instruction must be inserted before any flag setting, not only intentional, follows.

- forbidden are the following jumps: to itself, to the next instruction after jump, to the program entry, any conditional jump backwards that is not followed by flag setting before the jump, a jump forward behind a `RET` instruction that is not to the instruction next after the `RET`, unless other forward jump has already used that line number.

- more challenging rules govern the branching programs. The program generator maintains a status word for each branch. Statuses do split and join. The mentioned above requirement for the utilization of a set register is fulfilled, if at least one of the branches utilizes that register. On the other hand, the output vector must be set in each branch ending with a `RET`.

The PG frequently constructs exactly the same programs as the ones created in the past, which is intentional, for different concepts may contain the same code used in different contexts. A non-uniform distribution of the program generation output, however, is not welcomed. It is not the task of PG, but higher level phases, to decide which code would be more common eventually. To overcome this problem, a technique called hash-pooling has been applied.

First, an array of counters `count[N]` is initialized to 0 (N=$2^{14}$ in current implementation). Each time a new program `p` is created, an index (`h=hash(p) mod N`) is computed, where hash() is a non-cryptographic hash function. Let `T` be the total number of programs released so far. If `count[h]>2*(T/N)`, the program generation must be repeated. Otherwise, both `T` and `count[h]` are incremented by 1, and the created program is released. Consequently, no programs outputted by the PG are more frequent than twice the average.

Estimates have shown that, for the programs size limit of 7 instructions, the heuristic rules reduce the program search space from more than $10^{20}$ to as little as $10^{8}$, which is a substantial improvement over a purely random process. Though the heuristics slightly reduce the universality of the virtual machine, the resulting acceleration of the search in the programs space seems to be a substantial improvement over a pure universal search.

## 4. Evaluating the Concepts

The following discussion of reinforcement-learning and binary space partitioning is an abbreviated version of a more detailed presentation by Skaba (2012).

When a runtime execution is completed, the next action must be selected. Initially, the list of action-links is null. The list is filled-in during a process called exploration, discussed earlier. The space of all possible next actions is virtually unlimited. At any given time, however, the limit is set. Currently, the experimentally established limit is 50. In a continuous process of adding and removing the action-links, the number of actions that could ever be tested is also unlimited.

By action $a_i$ assigned to a given concept $A$ we mean a link to a program to be executed that is stored in the next concept $C$, the one $a_i$ points to, not the program stored in $A$. Should two different concepts *A* and *B* contain actions $a_i$ and $b_j$ linking to the same input of the same concept *C*, they would execute the same program (stored in *C*), but would have their own independent descriptions to be updated. In case of a multi-input concept, an action is executed only if all the inputs are available. In that case, a selection of an action reflects only the intension to execute an action, not its actual execution. Very likely, the action will never be executed, which is the case when a thread created by a `TERMINATED` runtime has timed-out before all the inputs were available.

The exploration probability is computed from the following equation:

$$P(Exploration) \propto \frac{Q_{const}}{Q_{const} + \sum_i Q_i}$$

where $Q_i$ is the current value of action $a_i$ and $Q_{const}$ is a predefined constant. With the increasing number of actions assigned to a concept and increasing value of those actions, the probability of exploration decreases. If, however, an exploration step is requested, and the number of actions assigned to a concept is equal to the mentioned above maximum limit, one of the actions—the one with the lowest value—will be replaced by the new action. Once a new action is added, its value is set to zero, and a temporary credit is assigned, in order that action is not removed until a more relevant value is established.

Likewise, the probability of selecting action $a_i$ (the exploitation) is governed by the equation:

$$P(a_i) = \frac{Q_i}{\sum_j Q_j}$$

## 4.1 Reinforcement Learning

The value $Q_{i,t}$ at time *t* is updated with the following TD-learning rule:

$$Q_{i,t} = Q_{i,t} + \alpha \left[ r_{i,t+1} + \gamma \overline{V}_{i,t+1} - Q_{i,t} \right]$$

where $r_{i,t+1}$ is the immediate reward at time *t+1*, $\alpha$ is the learning rate, $\gamma$ is the discount factor, and $\overline{V}_{i,t+1}$ is the mean value of all the actions of the concept the action $a_i$ points to, computed from the equation:

$$\overline{V}_t = \frac{\sum_j p_{j,t} Q_{j,t}}{\sum_j p_{j,t}}$$

where $p_{j,t}$ is the probability discussed in the next section.

## 4.2 Binary Space Partitioning

What matters in the listed above TD-learning update rule is the immediate reward $r_{i,t+1}$ that must be computed after the action execution has been completed. Leaving the discussion on the selection of the global fitness-function for later, we have to implement a sort of inductive bias that would let us evaluate concepts on the concept level, for the expected and the experimentally observed size of the concept hierarchy is to large to enable effective global-fitness propagation from the terminals to the roots.

The term *intrinsic motivation* was borrowed by the cognitive scientists from the psychology, to mean that an agent is engaged in an activity for its own sake, possibly for fun, rather than to fulfill some external drives. Closely related to it is the term *intrinsic reward* that controls the intrinsic motivation (Oudeyer and Kaplan, 2008).

Skipping the discussion on the relations between psychology and cognitive sciences, an information theory based approach is presented, that computes intrinsic reward and uses it as immediate reward for TD-learning. The proposed measure of intrinsic reward is based on the notion of *self-information* (Cover and Thomas, 1991) associated with the execution of a codelet, i.e., the amount of information provided by an event of a successful (i.e. pattern matching) program execution. The resulting intrinsic reward is an averaged value of self-information gain.

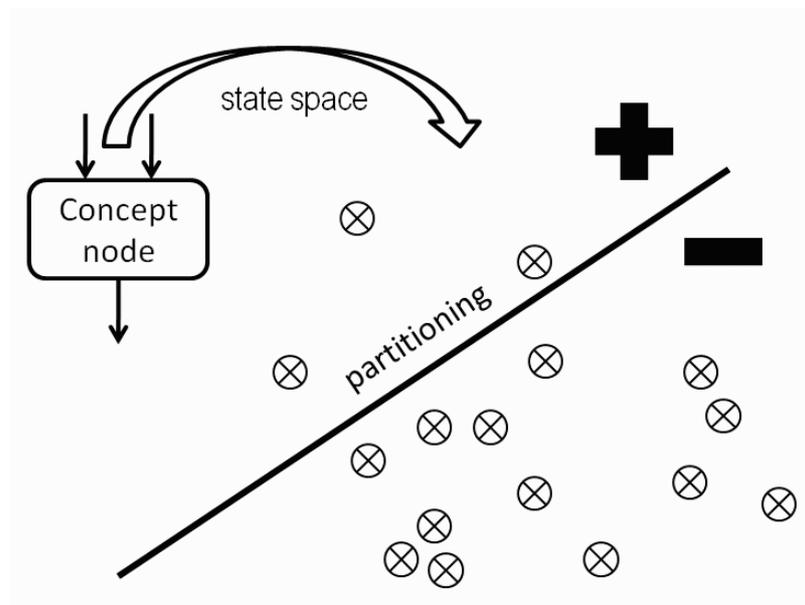

Figure 9. Partitioning of the state-space.

Figure 9 depicts the notion of binary space-partitioning. The input to a concept is a state space. It doesn't matter whether there is one or more inputs. All possible values of the input vectors constitute a multidimensional state space. The vectors of the state space are marked as crossed circles. A hyperplane, depicted as a line, divides the vectors to positive and negative examples.

Now imagine that $N > 0$ observed vectors consist of $N_{pos}$ positive examples and $N_{neg}$ negative examples, where $N = N_{pos} + N_{neg}$. Then, the probability of getting a positive example may be calculated from the equation:

$$p = \frac{N_{pos}}{N_{pos} + N_{neg}}$$

In case we are lucky to encounter a positive example, we will gain the self-information (in bits) given by the equation:

$$b = -\log_2 p$$

while, on average, we will be getting a reward:

$$r = -p \log_2 p$$

Since the input space is discrete, the defined above value of probability changes with every new example. While the total number $N$ increases, the rate of changes decreases.

Self-information may also be defined as a special case of Kullback-Leibler distance from a Kronecker delta representing the matching pattern to the probability distribution.

This approach has an interesting additive property, as well. If two concepts with probabilities $p$ and $q$ are executed in sequence, then the information gain is:

$$-\log_2 p - \log_2 q$$

The result we get is what would be expected, for if we integrate the codes of the two codelets, one partitioning the input space with probability $p$, the other with probability $q$, then we get the self-information measure of:

$$\log_2 (pq) = -\log_2 p - \log_2 q$$

Binary space partitioning is implemented by distinguishing two types of code execution termination: `RET` and `EXIT`, the first standing for a positive example, the second for a negative. Program generator restricts the usage of `EXIT` to cases where conditional branching occurs, while `RET` is always mandatory, like in the example shown on figure 10.

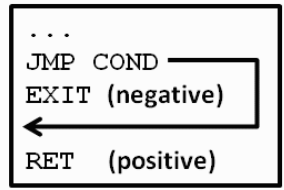

Figure 10. Machine code implementation of space partitioning.

An alternative definition of the meaning of a pattern may be concluded from the above. A pattern is any entity that is detectable by a space-partitioning conditional-jump of a codelet, that depends on the input only (not on the internal state), and effectively separates patterns from non-patterns. If a conditional jump is executed independently of the input data, the space is not partitioned at all. On the other hand, however, programs missing any `EXIT` instruction are legal and useful (`RET` is mandatory), too. We call them non-rewarding concepts. An example of a useful non-rewarding concept would be a program truncating input data, like a conversion of a color pixel into a grey-scale pixel.

Figure 11 shows the diagram for the $-p \log_2 p$ function, which looks like a distorted *binary entropy function*. It assumes 0.0 on both extremes. If one gets positive examples only, the concept is not rewarding at all. On the other extreme, if one gets positive examples rarely, they are individually most rewarding, but they are so rare, that—on average—the concept becomes marginally rewarding.

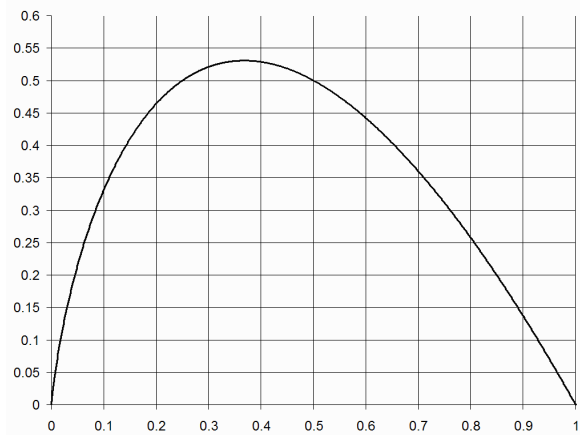

Figure 11. $-p \log_2 p$

This property may be illustrated with the following example. Imagine, we are expecting our cognitive engine to recognize all letters of the Latin alphabet. In one of possible approaches, we would expect the emergence of 25 individual concepts, each detecting the individual letters independently. Each of them would have rather low average reward, for the probabilities of the occurrence of each individual letter are rather low. We would also have to execute all of the concepts concurrently, consuming substantial computational resources.

An alternative approach would be based on, for example, checking whether the observed letter has some useful property (a feature), like contains a vertical bar. Some 14 Latin letters do have this property. Consequently, such a concept would divide the space of possible outcomes into two nearly equal groups, yielding high average reward. The process would be continued, involving other features. What follows, if two or more alternative approaches have evolved (as a result of self-programming) the cognitive engine would prefer the one that partitions the space more uniformly. The risk of expecting a very unique pattern is high and consequently very expensive.

### 4.3 Ultimate Goals and Sensory Feedback

There was little discussion on the global fitness function, so far, and no mention on how the actuators are evaluated and how the changes to the environment caused by the actuators, and the expected feedback from the sensory, is used to guide the learning process. These topics are a subject of the current study and experimentation. One of the implemented approaches is presented below.

The actuator-concepts—by definition—cannot be evaluated with the space partitioning approach, for they do not really partition the input space, and their implementation is not expressed in the code of the virtual machine. On the other hand, however, they act as terminal states and their values could potentially influence the learning process substantially.

The value $A$ of an actuator-concept could replace the $\overline{V}$ in the TD-learning rule (there is no space partitioning and no immediate reward) as follows:

$$Q_{i,t} = Q_{i,t} + \alpha \left[ \gamma A_{i,t+1} - Q_{i,t} \right]$$

First, however, we have to introduce the notion of *average reward per time step* (Sutton and Barto, 1998), here computed from real-time steps rather than discrete-time steps of MDP:

$$R_t = R_t + R_{t_0} e^{-\rho(t-t_0)}$$

where $R_t$ is the computed average reward at current real time $t$, $R_{t_0}$ is the average reward computed at time $t_0$ in the past (the last time it was updated), $r_t$ is immediate reward at the current time $t$, $\rho$ is a positive constant to control the rate of decay (to be set experimentally). The average reward is computed as a single value shared by all processes of the cognitive engine, updated every time an immediate reward is received. Following the computation of the average reward, the paradigm of maximizing the average reward as the global fitness function is proposed.

We can now put the following question: what is the impact of executing a given actuator concept on the overall average reward. For example, a movement of the robot's arm may cross the robot's visual field, or not. It will result in observing/detecting a pattern in the former case, and will not in the latter case. We can use the following rule to evaluate the actuator-concept:

$$A_{i,t} = A_{i,t} + \alpha [ \delta (R_t - R_{t_0}) - \mathrm{X}_{i,t_0} ]$$

where $t$ is the real-time of $A_{i,t}$ update, $t_0 < t$ is the real-time when the actuator-concept runtime was executed, $\alpha$ is the learning rate, $\delta > 0$ is a normalization coefficient, $\mathrm{X}_{i,t_o} > 0$ is the cost of executing the actuator-concept runtime at time $t_0$ and expressed in the same units as the reward. The value of $R_t - R_{t_0}$ may be negative. Even if $R_t - R_{t_0} = 0$, the value of $A_{i,t}$ will decrease. Should the value of an actuator-concept go below a predefined threshold, it would be deleted. A deletion of a given copy of actuator-concept does not exclude a successful application of other independent copy of actuator-concepts of the same template, possibly located in other locations of the concept hierarchy and executed with different values of the input parameters.

The actuator-concept value update algorithm works as follows. Once the actuator-concept runtime is executed, the current value $R_{t_0}$ is recorded, and the thread goes a-SLEEP, until time $t$ passes, when the value of $A_{i,t}$ is updated. A question arises, what time gap $t - t_0$ to select for the expected feedback from the sensory? For a physical robot we can expect the delays to be counted in tens or hundreds of milliseconds rather than nanoseconds. Since the time gap is quite large, the $\delta$ coefficient must be implemented as a function rather than a constant, to take into account the fact that multiple concurrently executed actuator-concepts may influence the average reward, and only a fraction of the average-reward volatility may be attributed to a given actuator-concept.

## 4.4    Artificial Economics

The idea of agent-based computational economics was applied in studies on AGI before (Goertzel, 2007). This section presents a brief presentation of how the notion of *artificial* *economics* and *complex adaptive system* was implemented to control the learning process of the AGINAO cognitive engine.

At the foundation of our approach there is an assumption that the operation of the cognitive engine must be performed within the available computational resources, the temporal resources especially, for the space (memory) restrictions do not matter so much. The concepts act as interacting adaptive agents, collaborating and/or competing, fighting for the computational resources and yielding the reward. A sample codelet, even if computing a function properly, will be discarded, if is too computationally expensive, or if not given an opportunity to estimate its cost, or if just useless for any other reason.

Every thread is assigned a limit of resources that decreases with every instruction executed on the virtual machine, and with any exploration/exploitation step. Should the resources be exhausted, the thread will be discarded, no matter what state it was in.

Once a pattern is detected, the thread is rewarded with extra resources that are proportional to the current self-information gain. Resources and rewards are expressed in the same units, subject to a normalization coefficient. A thread rewarded with more resources is capable of breeding more offspring (exploitation steps) and evolving more species (exploration steps).

Every thread is assigned a priority too, a positive number used to control the order of execution of the threads. The priority of a thread is set once at the time of thread creation and remains constant throughout thread's lifetime. In the current implementation, the priority is set as proportional to the value of the exploited action.

At every time, there is an overabundance of the threads awaiting in the priority queue, a phenomenon that could quickly lead to exponential growth. For that reason, a thread is assigned an expiration-time limit, too. Unlike the resources, that are not consumed while a thread is awaiting in the priority queue, the expiration-time is a real-time value that sets a deadline that is independent of any other processing, and eventually results in discarding the threads with the lowest priority.

Repetitive failures of attempts to execute a runtime will eventually cause a removal of the runtime-underlying concept, as well. Similar rules govern the expiration-times of the concepts and lead to concept deletion, should a concept rarely or never be executed as runtime.

A sort of equilibrium is maintained in the system. If, for example, the system load is low, the threads with lower absolute priority have greater chance to be executed. The system load depends on many factors, especially the flow of data coming from the sensory. Should the visual field be highly volatile, the cognitive engine would spend a lot of resources for sensory processing. Otherwise, the attentional focus would be directed towards tasks less dependent on the sensory. The exact design of the rules of economics is the question of current intensive study.

## 4.5    Concept Integration

An idea that has not been implemented yet is named *concept integration* (not to be confused with the integration of a concept into the hierarchy). Should two concepts be linked, provided some conditions are met, a new concept may be created based on the code of the linked concepts. The original concepts remain undisturbed. One may observe that this is a method of creating pieces of code much larger than what might be expected from the program-generator of the heuristic-search. An integrated concept may even be further integrated with other concepts, a process that leads to creation of programs virtually unlimited in size, yet useful. This process could possibly lead to converting the whole concept network to a single huge single concept program, though theoretically possible, doesn't seem to be useful in our case.

Should two concepts be integrated, we get two alternative methods or doing the same thing—one via the original (two) separate concepts, the other via the newly created concept. The RL that follows would determine whether either of the processing paths would be deleted.

Figure 12 presents how the integration could be used for creating a single-concept detector of the letter E, should the detection already be implemented with multiple concepts. A reader may notice that this example goes against the idea of partitioning the space uniformly, rather than uniquely. The truth is that it is the cognitive engine that decides which case is better. On the other hand, the integration may be very profitable for joining the non-rewarding concepts.

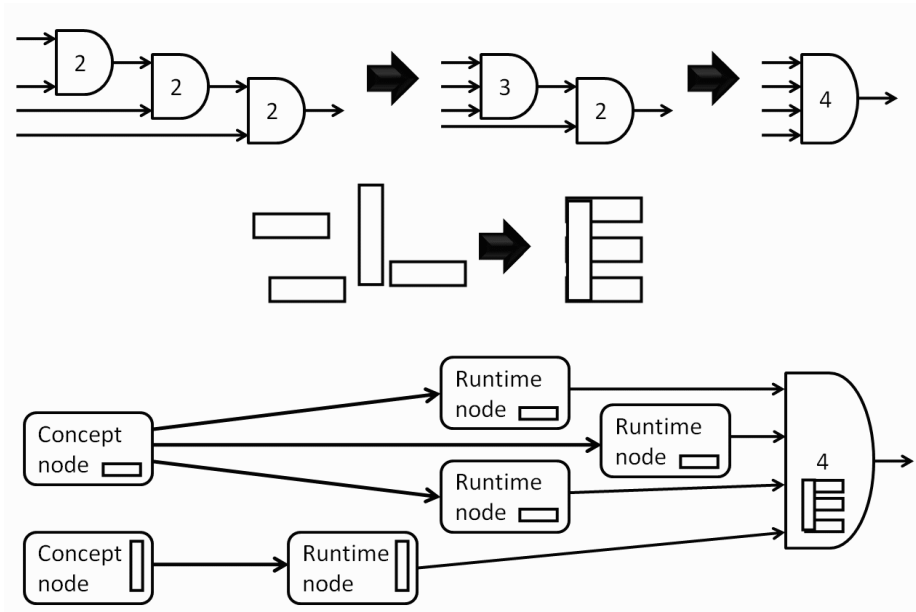

Figure 12. Concept integration.

## 5. Experimental results

Figures 13 shows a dump of the debugger that presents the operation of the virtual machine. In the upper part of the figure, the whole program is printed. Below, the two input vectors are shown that happen to be single-word strings with the same value (0008). The string size field is not depicted. In the lower part, the instructions are printed in the order as they were executed. To the right, the registers and their consecutive values are shown, each row for the value of the registers BEFORE the instruction in the same row was executed (see Figure 5 for reference). The registers are initialized to the default values, as show in row 1.

```
0000 MOV A,var1[00]
0005 CMP A,var2[00]
0010 APPEND A
0011 JNZ 0015
0014 RET
0015 EXIT

var1 = 0008
var2 = 0008
                        acc   idx   Z M MEM      OUT
0000 MOV A,var1[00]     0000 0000 N P 0000
0005 CMP A,var2[00]     0008 0000 N P 0000
0010 APPEND A           0008 0000 Z P 0000
0011 JNZ 0015           0008 0000 Z P 0000      0008
0014 RET                0008 0000 Z P 0000      0008
```

Figure 13. Program execution debugger dump.

The execution goes as follows. First, the value of `var1[0]` (0008) is moved to the accumulator. The result may be visible in the second row of the registers. The `CMP` instruction compares the value of the accumulator with `var2[0]`. Since the values are equal, the `ZERO` flag is set (`MINUS` flag is set too, but doesn't change), which is shown in row 3. Next, the value of the accumulator is outputted. The output, which was empty before, now is set to a single-word string with value 0008. Again, the size of the output is not depicted, as it may be concluded from the printout. The conditional jump is not executed, since the `ZERO` flag is set. Following, the `RET` instruction is performed, instead of the `EXIT` instruction, which would be executed should the `ZERO` flag be not set. The resulting output-string is the string of the last row. The `MEM` and `IDX` registers are not affected throughout the execution.

```
sources:2 outmax:2 memmax:2 exeestimate:7
0000 SIZE A,var1
0003 SUB A,var2[00]
0008 EXITM
0009 APPEND A
0010 RET
0000352[003]<2>-(000024[002]-2(0))(000007[001]-2(1)) cumul:  28015 value: 4667 selva:0
elapsed:0000.985 expires:1:084958.985 uses:    51260 axonsize:15 poatic[000000]
->0018001:1o---  d:  13  q:  45  r:   0 p:1000 b:   0 $: 5 e:0019.701 qmax:   0
->0018871:0o---  d: 131  q: 624  r:1362 p:  38 b: 4717 $: 5 e:0020.493 qmax: 624
->0000744:0---1  d:1000  q:5060  r:  47 p: 995 b:    7 $: 5 e:0002.611 qmax:6348
->0015295:0----  d: 333  q: 733  r: 292 p: 969 b:   45 $: 5 e:0019.376 qmax: 733
->0000786:0--t-  d: 596  q: 434  r:   0 p:   6 b: 7380 $: 0 e:0001.990 qmax: 434
->0000426:1o---  d:1000  q:6932  r:1106 p: 877 b:  189 $: 5 e:0001.506 qmax:8714
->0000524:0o---  d:1000  q:3456  r: 468 p: 950 b:   74 $: 5 e:0001.807 qmax:5608
->0010853:1----  d: 998  q: 874  r: 937 p:  25 b: 5321 $: 5 e:0015.313 qmax: 874
->0000805:0--t-  d:1000  q: 514  r:   0 p:  33 b: 4921 $: 0 e:0001.363 qmax:4369
->0000294:0----  d: 158  q:  20  r:   0 p:1000 b:   0 $: 6 e:0002.344 qmax:   0
->0000143:0----  d:1000  q:5071  r:1328 p: 850 b:  234 $: 5 e:0002.152 qmax:8239
->0018453:0o---  d: 133  q:  21  r:   0 p:1000 b:   0 $: 5 e:0021.056 qmax:   0
->0000662:0----  d:1000  q:4086  r:1541 p: 823 b:  281 $: 5 e:0001.848 qmax:7607
->0011755:0o---  d:  61  q:  31  r:   0 p:1000 b:   0 $: 5 e:0020.638 qmax:   0
->0013714:0o---  d:  11  q:  45  r:   0 p:1000 b:   0 $: 4 e:0031.651 qmax:   0
----------------------------------------
sources:2 outmax:1 memmax:1 exeestimate:7
0000 MOV A,var1[00]
0005 CMP A,var2[00]
0010 EXITZ
0011 APPEND A
0012 RET
0015295[012]<2>-(000352[003]-1(1))(003361[011]-1(1)) cumul:   500 value: 500 selva: 584
elapsed:0019.376 expires:0:002323.376 uses:      333 axonsize:2 poatic[000000]
->0029764:0o---  d:   5  q: 500  r:   0 p:1000 b:   0 $: 4 e:0032.548 qmax:   0
->0030389:1o---  d:  77  q:  38  r:   0 p:1000 b:   0 $: 6 e:0032.551 qmax:  43
```

Figure 14. Excerpt from the concept hierarchy.

Figure 14 shows an excerpt from the evolved concept-hierarchy that presents what cognitive structures are created. Two concepts, #352 and #15295, are depicted (two selected of possibly hundreds of thousands). For each concept, there is the code of the program that is followed by header information and by a list of actions (links) to other concepts.

From the header we can read that the concept #0000352 is at level [003] (sensory concepts are at level 0) and there are <2> inputs connecting from concepts: #000024 at level [002] and #000007 at level [001]. The level of a concept is defined as the highest level of the inputs plus 1. The list of next possible actions (one row per action) contains the concept number the link points to, followed by the number of the input the output is connected to. Letter 'o' means that a link is obsolete (invalid, to be removed), letter 't' denotes a link to an actuator-concept. The 'q:' column denotes the value of the link. One of the links (#0015295) points to the concepts shown below.

## Summary

This paper presented the details of the AGINAO self-programming engine, leaving away a more detailed discussion on the theory of mind underlying the project. The driving force of the task is an assumption that the NAO robot would be controlled within the limits of a contemporary (2012) desktop PC. The operation of the cognitive engine is conducted concurrently on many levels of the hierarchy, starting from the construction of simple random codelets consisting of instructions of a virtual machine, then sorted-out by a heuristic-search engine, then evaluated at runtime, and finally integrated into a hierarchy of dependencies. The structure of the concept-network is dynamic and open-ended. The model is tested on a physical NAO robot placed in a natural environment, in real time and concurrently with the process of building the concept-hierarchy. The learning is driven by a paradigm of maximizing the average reward per time step, that is measured with a purely information-theory based notion of self-information computed from binary space-partitioning.

## Appendix A. Instruction Set

The VM instruction set consists of 65 unique codes, not taking into account the parameter fields. The list below is given for reference only, it is incomplete, and it is not intended to be a detailed documentation. Some instructions, like temporal instructions, will not be found in typical microprocessors. The `A` stands for accumulator in mnemonic code notation, `var0` stands for output, `var1... varN` for N inputs. This section assumes some reader's knowledge of the mnemonic notation and typical processor's instructions.

- `MOV A,IDX;  MOV IDX,A;  ADD A,IDX;  SUB A,IDX;  CMP A,IDX;` the first argument is a destination register, excluding `CMP`. The `ADD, SUB` and `CMP` set flags. Similar rules apply to other instructions, respectively.

- `XCHG A,IDX;` exchange the contents of the registers. No flags are set.

- `MOVI A,int;     ADDI A,int;     SUBI A,int;     CMPI A,int;`
  `MOVI IDX,int;  ADDI IDX,int;  SUBI IDX,int;  CMPI IDX,int;` the `int` is an integer, encoded inline as the parameter field next to the instruction code.

- `MOVX A,varN[IDX];` `ADDX A,varN[IDX];` `SUBX A,varN[IDX];` `CMPX A,varN[IDX];` these instructions operate on the accumulator and the field of the `varN` vector pointed to by the current value of the `IDX` register. The result is stored in the accumulator. The flags are set. These instructions have a single inline encoded parameter, the value of N.

- `MOV A,varN[int];` `ADD A,varN[int];` `SUB A,varN[int];` `CMP A,varN[int];` first parameter is N, second parameter is the vector index, now encoded inline.

- `APPEND,A;` `SAVI [int],A;` `SAVX [IDX],A;` `SAV [int],int;` when the program execution is started, the size of the output (`var0`) is set to 0. `APPEND` appends the value of `A` to the current output and increases its current size by 1. `SAVI` and `SAVX` store `A` at position encoded as inline parameter or as current value of `IDX`, respectively. The `SAV` stores the value of the second inline parameter, rather than the accumulatoer.

- `ADDSAVI [int],A;` `ADDSAVX [idx],A;` `ADDSAV [int],int;` same as above, but the value is added to the current value, rather than stored.

- `MEMMOVI A,[int];` `MEMMOVX A,[IDX];` `MEMSAVI [IDX],A;` `MEMSAVX [IDX],A;` `MEMSAV [int],int;` same as above, but refers to local static memory.

- `INC A; DEC A; INC IDX; DEC IDX;` add/sub 1 and set flags.

- `NEG A;` Negation A = -A.

- `DELAY A,varN;` this is a temporal instruction that computes the difference (in milliseconds) between the time of creation of the executed runtime and the creation time of the runtime pointed-to by variable N, and stores the result in `A`.

- `WAIT A;` suspend execution for the number of milliseconds given in `A` and transit to `SLEEP` state.

- `SIZE A,varN; SIZE IDX,varN;` store the value of the size field of the variable N in the destination register.

- `FLAGS A; FLAGS IDX; FLAGS varN[IDX]; FLAGS varN[int];` set flags, according to the value of a register or a vector field.

- `RET; EXIT; EXITZ; EXITNZ; EXITM.` return control to the calling parent process, possibly the main execution loop. The difference between the return and the exit is discussed below. The three latter instructions exit conditionally, according to flag setting.

- `CALL;` launch another concept as a subroutine. Not yet implemented.

- `JMP ln; JZ ln; JNZ ln; JM ln;` the `ln` stands for the line number of the program code and is encoded as the first parameter. The first jump is unconditional, the other are conditional (jump on zero, jump on non-zero, jump on minus).